\documentclass{article}
\usepackage{spconf,amsmath,graphicx}

\usepackage{caption}
\usepackage{graphics}

\usepackage{parskip}

\usepackage{titlesec}
\titlespacing*{\section} {8pt}{8pt}{6pt}
\titlespacing*{\subsection} {4pt}{9pt}{4pt}
\titlespacing*{\subsubsection} {3pt}{9pt}{3pt}

\title{Robust saliency detection via fusing foreground and background priors}
\name{Kan Huang, Chunbiao Zhu* and Ge Li\thanks{This  work  was  supported  by  the  grant  of  National  Natural  Science  Foundationof China (No.U1611461),  the grant of Science and Technology  Planning  Project  of  Guangdong  Province,  China(No.2014B090910001) and the grant of Shenzhen PeacockPlan (No.20130408-183003656).}}
\address{School of Electronic and Computer Engineering Shenzhen Graduate School, Peking University \\ *zhuchunbiao@pku.edu.cn}
\begin{document}
%
\maketitle
\begin{abstract}
Automatic Salient object detection has received tremendous attention from research community and has been an increasingly important tool in many computer vision tasks. This paper proposes a novel bottom-up salient object detection framework which considers both foreground and background cues.
First, A series of background and foreground seeds are extracted from an image reliably, and then used for calculation of saliency map separately.
 Next, a combination of foreground and background saliency map is performed.
Last,  a refinement step based on geodesic distance is utilized to enhance salient regions, thus deriving the final saliency map.
Particularly we provide a robust scheme for seeds selection which contributes a lot to accuracy improvement in saliency detection.
Extensive experimental evaluations demonstrate the effectiveness of our proposed method against other outstanding methods.
\end{abstract}

\setlength{\parskip}{0em}

\section{Introduction}
\label{sec:intro}
The task of saliency detection is to identify the most important and informative part of a scene. It can be applied to numerous computer vision applications including image retrieval \cite{Hu2013Internet}, image compression\cite{Guo2010}, content-aware image editing\cite{Cheng2010RepFinder} and object recognition\cite{Ren2014Region}.Saliency detection methods in general can be categorized into
bottom-up models\cite{Achanta2009, Wang2008A} and top-down models\cite{Wang2014Salient}.
Bottom-up methods are data-driven, training-free, while top-down methods are task-driven, usually trained with annotated data.
Saliency models have been developed for eye fixation prediction and salient object detection.
Different from eye fixation prediction model which focus on identifying a few human fixation locations on natural images,
salient object detection model aims to pop out salient objects with well-defined boundaries, which is useful for many high-level vision tasks. In thin paper, we focus on the bottom-up salient object detection model.

\indent Research works that exploit foreground priors and explicitly extract salient objects from images have shown their effectiveness during past years. Contrast \cite{Achanta2009}\cite{Cheng2015Global} tends to be most influential factor among all kinds of saliency measures in exploiting foreground prior.
%
Some other works utilize the measure of rarity \cite{Ran2013What} to extract salient objects from images.
Suggested by Gestalt Psychological figure-ground assignment principle\cite{Mazza2005Foreground}, surroundedness cue has shown its effectiveness in saliency especially for eye fixation prediction, as in \cite{Zhang2016Exploiting} . But relying solely on it has trouble highlighting the whole salient objects.
Another effective type of salient object detection model is to exploit background priors in an image, from which salient objects could be detected implicitly.
By assuming most of the narrow border of the image as background regions, information of background priors can be exploited to calculating saliency map, as done in \cite{6751481,6619251}.
But it would also incur problems, for image elements distinctive to border regions are not always belonging to salient objects.

Unlike these methods that  focus either on exploiting foreground prior or on background prior for saliency detection, we endeavor to establish an efficient framework that integrate both foreground and background cues.
In this paper, a novel bottom-up framework is introduced. Surroundedness cue is utilized to exploit foreground priors, which are used for foreground seeds localization and subsequent saliency calculation. Meanwhile background priors are extracted from border regions and used for saliency calculation.
Two saliency maps based on background and foreground priors are generated separately, and then a combination is performed. Finally a refinement step based on geodesic distance is adopted to enhance the map and highlight the salient regions uniformly, deriving final saliency map.

Our work has several contributions: (1) surroundedeness cue is utilized to exploit foreground priors, which is proved to be effective for saliency detection when combined with foreground prior; (2) a robust seed estimation scheme which contributes a lot to accuracy of saliency detection is established; (3) a framework that integrate both foreground and background priors is proposed.

\section{Proposed approach}
\label{sec:proposedapproach}

\indent There are two main paralleled sub-processes in our framework: background saliency and foreground saliency. They are calculated separately based on foreground and background seeds. Then two saliency maps are fused into one and enhanced by a refinement step based on geodesic distance to derive the final saliency map. The main framework of proposed approach is depicted in Fig.1.

\begin{figure}[htb]
\setlength{\abovecaptionskip}{0.2cm}
\setlength{\belowcaptionskip}{-0.cm}
\centering
\includegraphics[width = 0.46\textwidth]{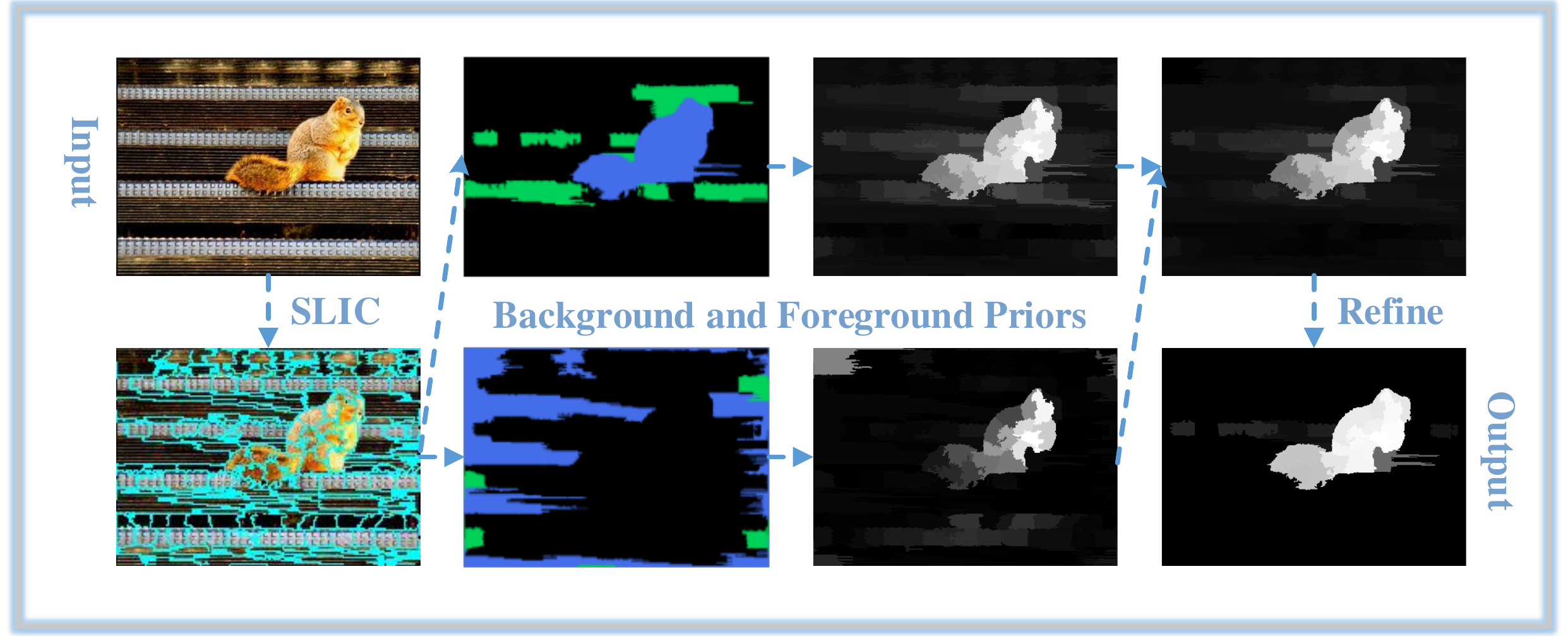}
\captionsetup{justification=centering}
\caption{Overview of the main framework of our proposed approach}
\end{figure}

\subsection{Foreground saliency}
This section will detail on how to find reliable foreground seeds and generate saliency map based on these selected seeds.
\subsubsection{Foreground seeds estimation}
To extract foreground seeds from an image reliably, surroundedness cue is employed. 
We adopt the binary segmentation based method in BMS\cite{Zhang2016Exploiting} , which exploit surroundedness cue thoroughly in an image, to guide our foreground seeds localization.
We denote the map generated by BMS as a surroundedness map, $S_{B}$, in which pixel value indicates its degree of surroundedness.
To better utilize structural information and abstract small noises, We decompose image into a set of superpixels by SLIC algorithm \cite{Achanta2012SLIC}. All operation in rest of this paper is performed on superpixel-level. The surroundedness value of each superpixel is defined by averaging the value $S_{B}$ of all its pixels inside, denoted by $S_p(i), i=1,2,...,N$. $N$ is the number of superpixels.

Unlike previous works\cite{6751481,6619251} that treat some regions as certain seeds, we provide a more flexible scheme for seeds estimation. We define two types of seed elements: strong seeds and weak seeds. Strong seeds have high probability of belonging to foreground/background while weak seeds have relatively low probability of belonging to foreground/background. For foreground seeds, the two types of seeds are selected by:
\begin{equation}
\label{fore1}
C_{fore}^+ = \{  i | S_{p}(i) >= 2 \cdot mean(S_{p}) \}
\end{equation}
\begin{equation}
\label{fore2}
\begin{aligned}
C_{fore}^- = \{  i | &S_{p}(i) >= mean(S_{p}) ~~and~~ \\
&S_{p}(i) < 2 \cdot mean(S_{p}) \}
\end{aligned}
\end{equation}
where $C^+$ denotes the set of strong seeds and $C^-$ weak seeds,
$i$ represent $i$th superpixel. mean(.) is the averaging function. It is obvious from formula (\ref{fore1})(\ref{fore2}) that elements of higher degree of surroundedness are more likely to be chosen as strong foreground seeds, which is consistent with intuition.


\subsubsection{Foreground saliency map}
For saliency calculation based on given seeds, a ranking method in \cite{Zhou2003Ranking} that exploits the intrinsic manifold structure of data for graph labelling is utilized. The ranking method is to rank the relevance of every element to the given set of seeds.
We construct a graph that can represent an whole image as in work \cite{Yang2013Saliency}, where each node is a superpixel generated by SLIC.
%

The ranking procedure is as follows: Given a graph $G = (V, E)$ ,where the nodes are $V$ and the edges $E$ are weighted by an affinity matrix $\boldsymbol{W} = [w_{ij}]_{n \times n}$. The degree matrix is defined by $\boldsymbol{D} = diag\{  d_{11}, ... , d_{nn} \}$, where $d_{ii} = \sum_j w_{ij}$.
The ranking function is given by:
\begin{equation}
\boldsymbol{g}^* = ( \boldsymbol{D} - {\alpha}{\boldsymbol{W}} )^{-1} \boldsymbol{y}  \label{eq}
\end{equation}
The $\boldsymbol{g}^*$ is the resulting vector which stores the ranking results of each element. The $\boldsymbol{y} = [y_1, y_2, ... ,y_n]^{T}$ is a vector indicating the seed queries.

 In this work, the weight between two nodes is defined by:
\begin{equation}
w_{ij} = e^{-\frac{\| c_i - c_j \|}{ {\sigma}^2 }}
\end{equation}
where $c_i$ and $c_j$ denote the mean of the superpixels corresponding to two nodes in the CIE LAB color space, and $\sigma$ is a constant that controls the strength of the weight.

Different from \cite{Yang2013Saliency} that define $y_i = 1$ if $i$ is a query and $y_i = 0$ otherwise, we define $y_i$ as the strength of the query extra. That is, $y_i = 1$ if $i$ is a strong query, and $y_i = 0.5$ if $i$ is a weak query, and $y_i = 0$ otherwise.

For foreground seeds based ranking, all elements are ranked by formula (4) given the sets of seeds in (1)(2). The process of foreground saliency is illustrated in Fig.2(first row).


\begin{figure}[htb]
\setlength{\abovecaptionskip}{-0.4cm}
\setlength{\belowcaptionskip}{-0.cm}
\centering
\includegraphics[width = 0.48\textwidth]{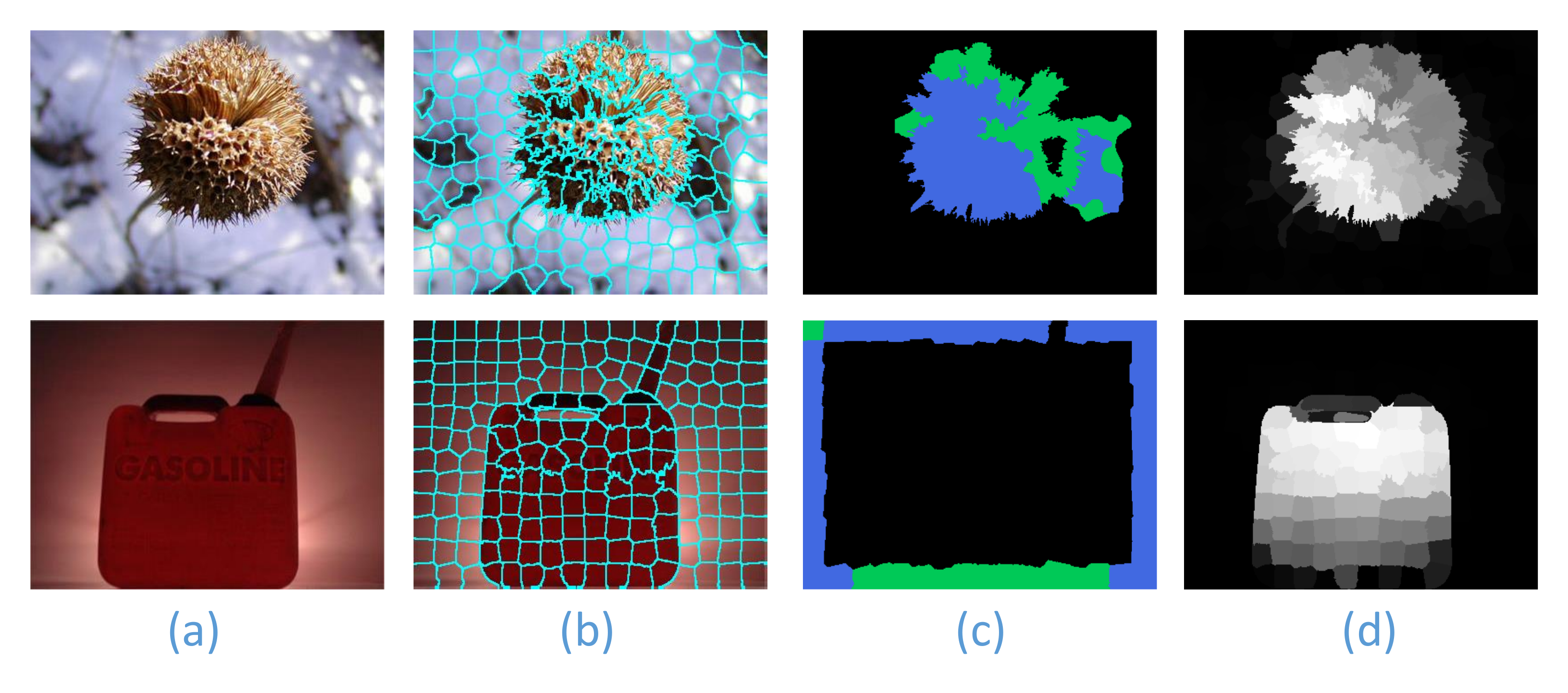}
\captionsetup{justification=centering}
\caption{Illustration of foreground and background saliency. (a) original image; (b) superpixel segmentation; (c)top: foreground seeds, bottom: background seeds(blue : mask of strong seeds, green: mask of weak seeds); (d)top: foreground saliency map, bottom: background saliency map.}
\end{figure}

\subsection{Background saliency}
Complementary to foreground saliency, background saliency aims to extract regions that are different from background in feature distribution.
We first select a set of background seeds and then calculate saliency of every image element according to its relevance to these seeds.
This section elaborates on the process of seeds estimation and background saliency calculation.

\begin{figure}[htb]
\setlength{\abovecaptionskip}{0cm}
\setlength{\belowcaptionskip}{-0.cm}
\centering
\includegraphics[width = 0.46\textwidth]{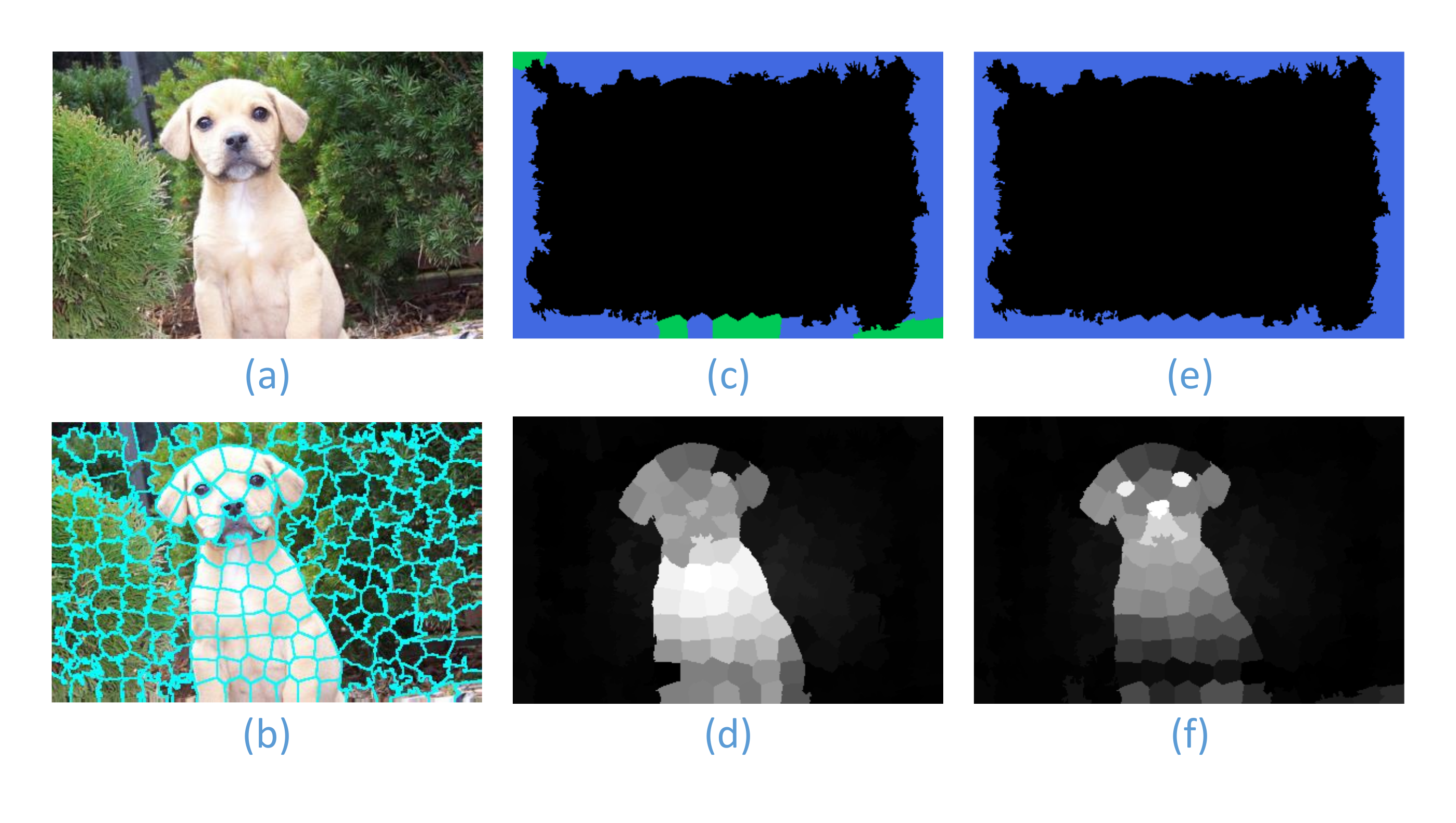}
\captionsetup{justification=centering}
\caption{Comparison of different seeds estimation scheme. (a) original image; (b) superpixel segmentation; (c) our scheme for background seeds estimation; (d) saliency map corresponding to (c); (e) common seeds estimation scheme; (f) saliency map corresponding to (e).}
\end{figure}

\subsubsection{Background seeds estimation}
Unlike most previous works \cite{Wang2014Salient} that use the elements on image boundary as background seeds, we divide the elements on image border into two categories(strong seeds and weak seeds) as in foreground situation.
%
We denote the average value of all border elements as $\overline{c}$.
The euclidean distance between each feature vector and the average feature vector is computed by ${\bf{d}}_c = \bf{dist} (c, \overline{c})$,
 the average of ${\bf{d}}_c $ is denoted by $\overline{{\bf{d}}_c}$. The background seeds are estimated by:

\begin{equation}
C_{back}^+ = \{  i | {\bf{d}}_c (i)  >= 2 \cdot\overline{{\bf{d}}_c} \}
\end{equation}
\begin{equation}
C_{back}^- = \{  i | {\bf{d}}_c (i)  >= \overline{{\bf{d}}_c} ~~ and~~  {\bf{d}}_c (i)  < 2 \cdot \overline{{\bf{d}}_c}  \}
\end{equation}
where $C_{back}^+$ denotes strong background seeds, $C_{back}^-$ denotes weak background seeds.

\subsubsection{Background saliency map}
Similar to foreground situation, the value of indication vector for background seeds $\boldsymbol{y}$ is $y_i = 1$ if $i$ belongs to $C_{back}^+$,
$y_i = 0.5$ if $i$ belongs to $C_{back}^-$
and 0 otherwise.
Relevance of each element to background seeds is computed by formula (\ref{eq}).
Elements in resulting vector ${\boldsymbol{g}}^*$ indicates the relevance of a node to the background queries, and its complement is the saliency measure.The saliency map using these background seeds can be written as:
\begin{equation}
S(i) = 1 - {\boldsymbol{g}}^*(i)   \qquad i=1,2,...,N.
\end{equation}

The process of background saliency is shown in Fig.2(second row), and comparison between our seeds estimation scheme and common scheme is illustrated in Fig.3. It is noted that our scheme is robust for extracting more salient regions from an image.

\begin{figure*}[!bth]
\setlength{\abovecaptionskip}{0cm}
\setlength{\belowcaptionskip}{-0.cm}
\label{comparison}
\centering
\includegraphics[width = 0.99\textwidth]{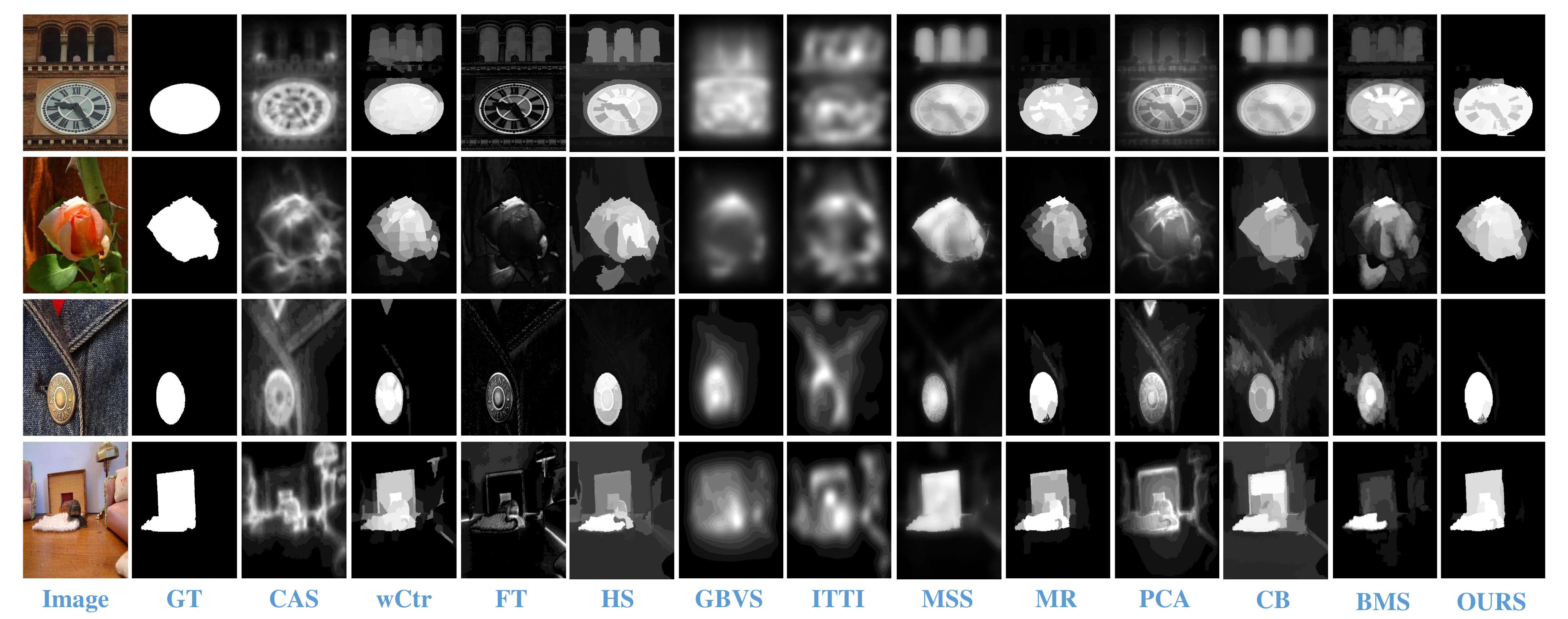}
\captionsetup{justification=centering}
\caption{Visual comparison of saliency models}
\end{figure*}

\begin{figure*} [!htb]
\setlength{\abovecaptionskip}{0.cm}
\setlength{\belowcaptionskip}{-0.cm}
\centering
\includegraphics[width=1.00\textwidth]{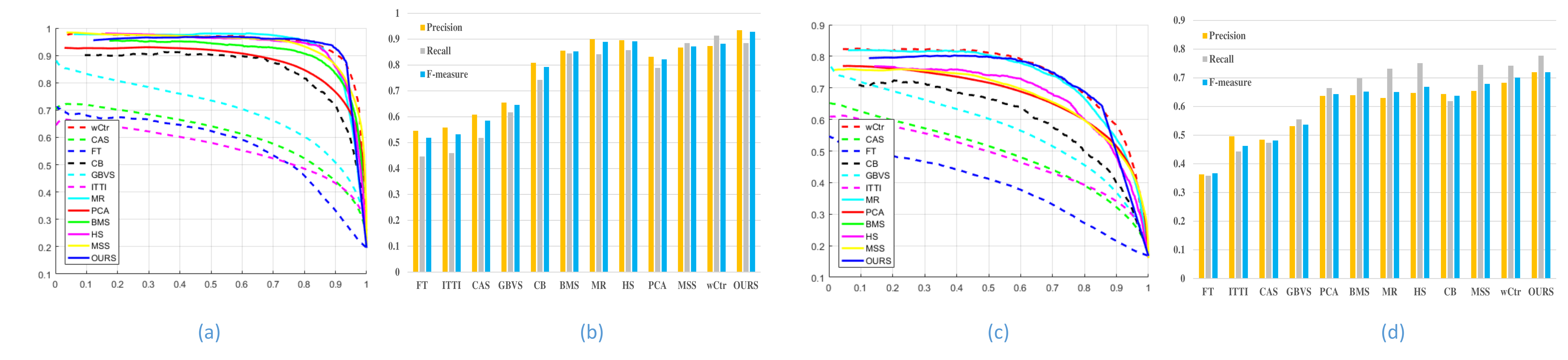}
\caption{(a) PR curve on ASD dataset; (b) precision, recall and F-measure on ASD dataset; (c) PR curve on DUT-OMRON dataset; (d) precision, recall and F-measure on DUT-OMRON dataset; }
\label{fig:pictures2}
\end{figure*}

\begin{table*}[!htb]
\setlength{\abovecaptionskip}{0.cm}
\setlength{\belowcaptionskip}{-0.cm}
\caption{Quantitative comparision of MAE and AUC on ASD dataset}
\centering
\begin{tabular}{|c|c|c|c|c|c|c|c|c|c|c|c|c|}
\hline
Method & BMS     & CAS     & CB    & FT           & GBVS    & ITTI      & HS      & MR       & PCA     & wCtr & MSS  & OURS\\
\hline
MAE      & 0.1105 & 0.2293 & 0.1707 & 0.2195 & 0.2125 & 0.2464 & 0.1123  &         0.0743 & 0.1556   &   0.0689  & 0.1045 & \textbf{0.0596} \\
\hline
AUC       & 0.8990 & 0.5903 & 0.8412 & 0.4980 & 0.6754  & 0.5335 &  0.9303 &         0.9308   &  0.8772 & 0.9390  & 0.9344 &   \textbf{0.9446}    \\
\hline
\end{tabular}
\end{table*}

\begin{table*}[!htb]
\setlength{\abovecaptionskip}{0.cm}
\setlength{\belowcaptionskip}{-0.cm}
\caption{Quantitative comparision of MAE and AUC on DUT-OMRON dataset}
\centering
\begin{tabular}{|c|c|c|c|c|c|c|c|c|c|c|c|c|}
\hline
Method & BMS     & CAS     & CB    & FT       & GBVS   & ITTI     & HS           & MR      & PCA     & wCtr      & MSS       & OURS\\
\hline
MAE     & 0.1485 & 0.2406 & 0.1262  & 0.2363 & 0.2278 & 0.2499 & 0.1798  & 0.1418  & 0.1843  & 0.1143 &  0.1646  &  \textbf{0.1068} \\
\hline
AUC      & 0.6676  & 0.4674 & 0.6648 & 0.3255 & 0.5330 & 0.4450 &  0.7010  &  0.7259   & 0.6700  & \textbf{0.7846} &  0.7202  &0.7619 \\
\hline
\end{tabular}
\end{table*}

\subsection{Geodesic distance refinement}
 A combination of Foreground  and background saliency maps is performed as follows: elements whose value is larger than the average value of that map is selected as saliency elements separately in these two maps and combined into one set, a ranking is conducted again using these elements as seeds to get a combination map $S_{com}$.

The final step of our proposed approach is refinement with geodesic distance \cite{Zhu2014Saliency}. The motivation underlying this operation is based on observation that determining saliency of an element as weighted sum of saliency of its surrounding elements, where weights are corresponding to Euclidean distance, has a limited performance in uniformly highlighting salient object. We tend to find a solution that could enhance regions of salient object more uniformly. From recent works \cite{Fu2014Geodesic} we found the weights may be sensitive to
geodesic distance.

For $j$th superpixel, its posterior probability can be denoted
$S_{com}(j)$, thus the saliency value of the $q$th superpixel is refined by geodesic distance as follows:
\begin{equation}
S_{final}(q) = \sum_{j=1}^{N} {\delta}_{qj}\cdot {S_{com}}(j)
\end{equation}
where $N$ is the total number of superpixels, and ${\delta}_{qj}$ is a weight based on geodesic distance \cite{Zhu2014Saliency} between $q$th and $j$th superpixel.
Based on the graph model constructed in section 2.1.2 , the geodesic distance between two superpixels $d_g(p, i)$ can be defined as accumulated edge weights along their shortest path on the graph:
\begin{equation}
d_g(p,i) = \min_{a_1 = p, a_2, a_3,..., a_n = i} \sum_{k=1}^{n-1} d_c(a_k, a_{k+1})
\end{equation}
In this way we can get geodesic distance between any two superpixels in the image. Then the weight $\delta_{pi}$ is defined as
$
\delta_{qj} = exp\lbrace - \frac{d_{g}^{2}(p,i)}{2\sigma_c^{2}} \rbrace
$
where $\sigma_c$ is the deviation for all $d_c$ values.
The salient objects are highlighted uniformly after this step of processing, as will be seen in experiment section.

%
%
%

\section{Experiment}
\label{sec:exp}
This section presents evaluation of our proposed method.

{\bf{Datesets.}}
We test our proposed model on ASD dataset \cite{Achanta2009}, OUT-OMRON dataset \cite{Yang2013Saliency}. ASD dataset provides 1000 images with annotated object-contour-based ground truth, while DUT-OMRON dataset provide 5168 more challenging images with pixel-level annotation.

{\bf{Evaluation metircs.}}
For accurate evaluation, we adopts four metrics: Precion-recall(PR) curve, F-measure, mean absolute error(MAE), and AUC score.
Fig.5 shows the PR curves, and precision, recall and F-measure values for adaptive threshold that is defined as twice the mean saliency of the image. Table 1 and table 2 shows the MAE and AUC scores on two datasets.




{\bf{Comparison}}
We compare our proposed method with 11 state-of-the-art models, including CAS\cite{Goferman2012Context}, wCtr\cite{Zhu2014Saliency}, FT\cite{Achanta2009}, DRFI\cite{Wang2014Salient}, GBVS\cite{Sch2007Graph}, ITTI\cite{730558}, MILPS\cite{Huang2017Salient}, MR\cite{6619251}, PCA\cite{Ran2013What}, SBD\cite{Zhao2016Saliency}, BMS\cite{Zhang2016Exploiting} . It is noted that  our method highlights salient regions more uniformly and achieves better results especially in PR curve, MAE scores. In general our method outperforms other competitive approaches.

\section{conclusion}
In this paper, we present a novel and efficient framework for salient object detection via complementary combination of foreground and background priors. The key contributions of our method are: (1) surroundedness cue is utilized for exploiting foreground prior, which is proved to be extremely effective when combined with backgournd prior.
(2) A robust seed estimation scheme is provided for seeds selection with their confidence of belonging to background/foreground estimated. Extensive experimental results demonstrate the superiority of our proposed method against other outstanding methods. Our proposed also has a efficient implementation which is useful for real-time applications.

\setlength{\parskip}{0.3em}
\bibliographystyle{IEEEbib}
\bibliography{strings,refs}

\end{document}